\pdfoutput=1
\documentclass[11pt]{article}

\usepackage[]{EMNLP2023}

\usepackage{times}
\usepackage{latexsym}
\usepackage{amsmath}
\usepackage{amsfonts}

\usepackage[T1]{fontenc}
\usepackage[utf8]{inputenc}

\usepackage{microtype}

\usepackage{inconsolata}

\usepackage{braket}
\usepackage{verbatim}
\usepackage{ulem}
\usepackage{adjustbox}
\usepackage{multirow}
\usepackage{enumitem}

\usepackage[percent]{overpic} % positions via \put(x%,y%)
\usepackage{xcolor}
\usepackage{hyperref}
\usepackage{booktabs}
\newcommand{\na}{--}
\newcommand{\fqok}[1]{\textbf{#1}$^{\checkmark}$} % fq > 0.1
\newcommand{\cell}[2]{#1, #2}                      % utility, forget quality

\title{Evaluating Cross-Lingual Unlearning in Multilingual Language Models}
\author{ Tyler Lizzo, Larry Heck\\ AI Virtual Assistant (AVA) Lab\\
   Georgia Institute of Technology  \\ \texttt{\{lizzo,larryheck\}@gatech.edu}}

\begin{document}
\maketitle
\setlength{\belowcaptionskip}{-10pt}

\begin{abstract}
We present the first comprehensive evaluation of cross-lingual unlearning in multilingual LLMs. Using translated TOFU benchmarks in seven language/script variants, we test major unlearning algorithms and show that most fail to remove facts outside the training language, even when utility remains high. However, subspace-projection consistently outperforms the other methods, achieving strong cross-lingual forgetting with minimal degradation. Analysis of learned task subspaces reveals a shared interlingua structure: removing this shared subspace harms all languages, while removing language-specific components selectively affects one. These results demonstrate that multilingual forgetting depends on geometry in weight space, motivating subspace-based approaches for future unlearning systems.
\end{abstract}

\section{Introduction}

Large language models internalize vast amounts of factual and contextual information during pretraining, including personal data, copyrighted material, and potentially harmful or toxic content. This has made unlearning an increasingly important capability for compliance with regulations such as GDPR, for removal of proprietary data, and for mitigating safety risks in deployed systems. A number of unlearning techniques have been proposed to address this, spanning from architectural to data-driven and parameter-based methods, yet almost all have been developed and evaluated exclusively in monolingual settings.

Cross-lingual unlearning introduces challenges that do not arise in monolingual settings. LLMs such as mBERT \cite{wu-dredze-2019-beto}, Llama \cite{grattafiori2024llama3herdmodels}, and Mixtral \cite{jiang2024mixtralexperts} are increasingly trained as multilingual systems, and they tend to encode semantically aligned representations across languages through shared vocabularies and multilingual training objectives. Prior work in multilingual representation learning \cite{wu-dredze-2019-beto} suggests that these models form partially shared semantic subspaces, often described as an interlingua-like structure that enables zero-shot transfer. When factual knowledge occupies overlapping parameter subspaces across languages, removing it in one language may not eliminate its representation in others and may unintentionally degrade broader multilingual capability. This interdependence makes multilingual unlearning more complex and highlights the need for approaches that operate on shared geometric structure rather than relying only on language-specific loss signals.

This paper conducts a systematic evaluation of cross-lingual unlearning across a broad range of algorithms, including gradient ascent \cite{jang2023knowledge}, KL-based
objectives\cite{wang2025balancing}, gradient-difference variants \cite{liu2022continual}, preference
optimization \cite{rafailov2024directpreferenceoptimizationlanguage}, FLAT \citep{wang2024flat}, NPO \citep{zhang2024npo}, and a subspace-projection method UNLEARN \citep{lizzo2025unlearn}. Across models and languages, we find that most existing techniques struggle to remove targeted knowledge outside the ``unlearned" language or introduce substantial collateral degradation across languages.

Among the methods evaluated, the subspace-projection approach exhibits noticeably stronger cross-lingual forgetting. Because it operates directly in weight space by identifying and removing task-specific subspaces, it generalizes effectively across languages and preserves overall model utility. We analyze this behavior through several controlled studies, including unlearning a fact in a single language and measuring retention of that fact in other languages. Because subspace-projection methods identifies the subspace of a domain, we extend this to identify an interlingua space, a language-independent representation within the model. This enables further analysis into cross-lingual unlearning, both when we want to unlearn information across languages and when we want to isolate the unlearning process to a single language.

The contributions of this work are: 1) A systematic evaluation of cross-lingual unlearning across a diverse set of methods and multilingual model families using translated variants of the TOFU benchmark \cite{maini2024tofutaskfictitiousunlearning}.
2) Analysis that reveals most existing unlearning algorithms fail to generalize forgetting across languages and often introduce collateral degradation, revealing fundamental limitations in current approaches. 3) Experimental results and analysis that show the subspace-projection approach achieves substantially stronger cross-lingual unlearning by representing an interlingua subspace to capture language-independent structure, enabling new forms of controlled multilingual forgetting.

This paper investigates cross-lingual unlearning in multilingual language models and evaluates a wide range of unlearning methods. Section \ref{relatedworks} reviews prior work in unlearning and multilingual representation learning. Section \ref{experiments} introduces the experimental setup and outlines the three evaluation settings: one-to-one unlearning, many-to-one unlearning, and transliteration experiments, and presents results across model families and methods. Section \ref{discussion} discusses the uncovered interlingua subspace and its role in multilingual forgetting. Section \ref{conclusions} concludes with a summary of our findings and future research directions.

\section{Related Works}
\label{relatedworks}

\subsection{Unlearning Methods}
Unlearning techniques for large language models can be grouped into architecture centric, data centric, parameter centric, and hybrid approaches. Architecture centric methods modify or augment the model structure, such as through adapters or contrastive decoding assistants, to suppress forgotten content at inference time. These approaches are modular and often reversible, but they typically leave internal representations intact and are not designed for cross-lingual forgetting.

Data centric approaches modify the learning signal. A classical example is gradient ascent \citep{jang2023knowledge}, which maximizes loss on the forget set. Stabilized variants incorporate KL based retain objectives \citep{wang2025balancing}, gradient difference corrections \citep{liu2022continual}, or f divergence maximization as in FLAT \citep{wang2024flat}. Preference-based methods recast forgetting as preference optimization: PO \citep{rafailov2024directpreferenceoptimizationlanguage} and its unlearning variant NPO \citep{zhang2024npo} encourage the model to downweight responses associated with the forget data. These methods can be effective in monolingual settings but rely heavily on surface form cues, which as we later show limits their generalization across languages.

Parameter centric methods intervene directly in weight space. Task vector subtraction \citep{ilharco2023editing} and related update negation strategies attempt to remove directions associated with unwanted behaviors. The most structured approach is UNLEARN \citep{lizzo2025unlearn,lizzo2025patent}, a subspace-projection method that identifies a low dimensional subspace corresponding to the forget task and removes it through projection. Through subspace discrimination, UNLEARN separates overlapping components of forget and retain tasks, enabling targeted removal with minimal collateral degradation. 
%As our results demonstrate, this geometric perspective provides substantially better cross-lingual forgetting than loss based methods.

Hybrid and system level techniques such as SISA style retraining \citep{kadhe2023fairsisaensemblepostprocessingimprove} and RAG based filtering \cite{wang2024machineunlearningmeetsretrievalaugmented} combine elements of multiple categories or operate outside the model. These approaches can be practical but do not alter internal representations and therefore do not address the challenges of cross-lingual unlearning.

\subsection{Multilingual Representation Learning}
Multilingual language models exhibit strong cross-lingual generalization, suggesting that they encode partially shared semantic representations across languages. Early evidence comes from mBERT, which supports zero shot transfer on downstream tasks despite never being trained with parallel data \citep{pires-etal-2019-multilingual,wu-dredze-2019-beto}. Subsequent models such as XLM R \citep{conneau2020unsupervisedcrosslingualrepresentationlearning} strengthen this finding. Sentence level encoders provide even clearer evidence of shared structure: LASER \citep{artetxe-schwenk-2019-massively} and LaBSE \citep{feng2022languageagnosticbertsentenceembedding} learn language agnostic embeddings that align semantically equivalent sentences across more than 100 languages. These results indicate that multilingual pretraining induces a partially-shared, interlingua-like representation space.

This shared space is structured rather than uniform. Probing studies show that multilingual encoders capture common syntactic and semantic abstractions, but alignment quality varies with linguistic similarity, tokenization overlap, and training signal \citep{pires-etal-2019-multilingual,k2020crosslingualabilitymultilingualbert}. Analyses of multilingual representation geometry further show that languages overlap substantially in higher level semantic subspaces while retaining language specific components \citep{chi-etal-2020-finding}. Thus, multilingual language models encode knowledge in a space that is partly unified across languages but meaningfully differentiated across language families.

\begin{comment}
I think this paragraph is a bit redundant and out of place here (this subsection is on multilingual representation learning, not unlearning. I moved this to the next section.
These representational properties directly impact unlearning. If factual knowledge expressed in different languages corresponds to overlapping parameter subspaces, forgetting in one language may transfer to others. However, incomplete alignment can produce asymmetric forgetting where removal succeeds in the source language but leaves residual traces elsewhere. This motivates unlearning approaches that operate directly in weight space or learned task subspaces rather than relying solely on surface level loss signals.
\end{comment}
% This starts to oversell UNLEARN. 
% As we demonstrate, subspace based methods such as UNLEARN are well suited to this setting.

\subsection{Cross-Lingual Unlearning}
Despite interest in model unlearning, cross-lingual unlearning remains largely unexplored. Early work examines whether removing knowledge in one language transfers to others. Choi et al. \citep{choi2024crosslingualunlearningselectiveknowledge} introduce language specific adapters to suppress facts across languages, and Lu et al. \citep{lu2025learnunlearnaddressingmisinformation} evaluate multilingual forgetting of misinformation and observe degradation in unrelated languages. Translation based pipelines, which forget in English and evaluate in other languages, typically show limited transfer and inconsistent forgetting behavior.

The representational properties of multilingual learning directly impact unlearning. If factual knowledge expressed in different languages corresponds to overlapping parameter subspaces, forgetting in one language may transfer to others. However, incomplete alignment can produce asymmetric forgetting where removal succeeds in the source language but leaves residual traces elsewhere. This motivates unlearning approaches that operate directly in weight space or learned task subspaces rather than relying solely on surface level loss signals.

\begin{comment}
Shouldn't this paragraph come after the Experiments - you're making claims that are not yet supported in the literture but rather supported by the experiments in this paper.

    Existing methods lack mechanisms for targeting multilingual semantic representations and remain sensitive to language specific surface forms. Prior studies report that monolingual unlearning often fails to generalize cross linguistically and can introduce collateral damage in uninvolved languages. This motivates a principled examination of how forgetting propagates through shared multilingual subspaces and highlights the need for weight space approaches designed to operate on language agnostic structure. As we show, subspace based methods such as UNLEARN provide a more reliable foundation for cross-lingual unlearning.
\end{comment}

\section{Experiments and Results}
\label{experiments}

\subsection{Experimental Setup and Metrics}

\begin{figure}
    \centering
  \begin{overpic}[width=0.95\linewidth]{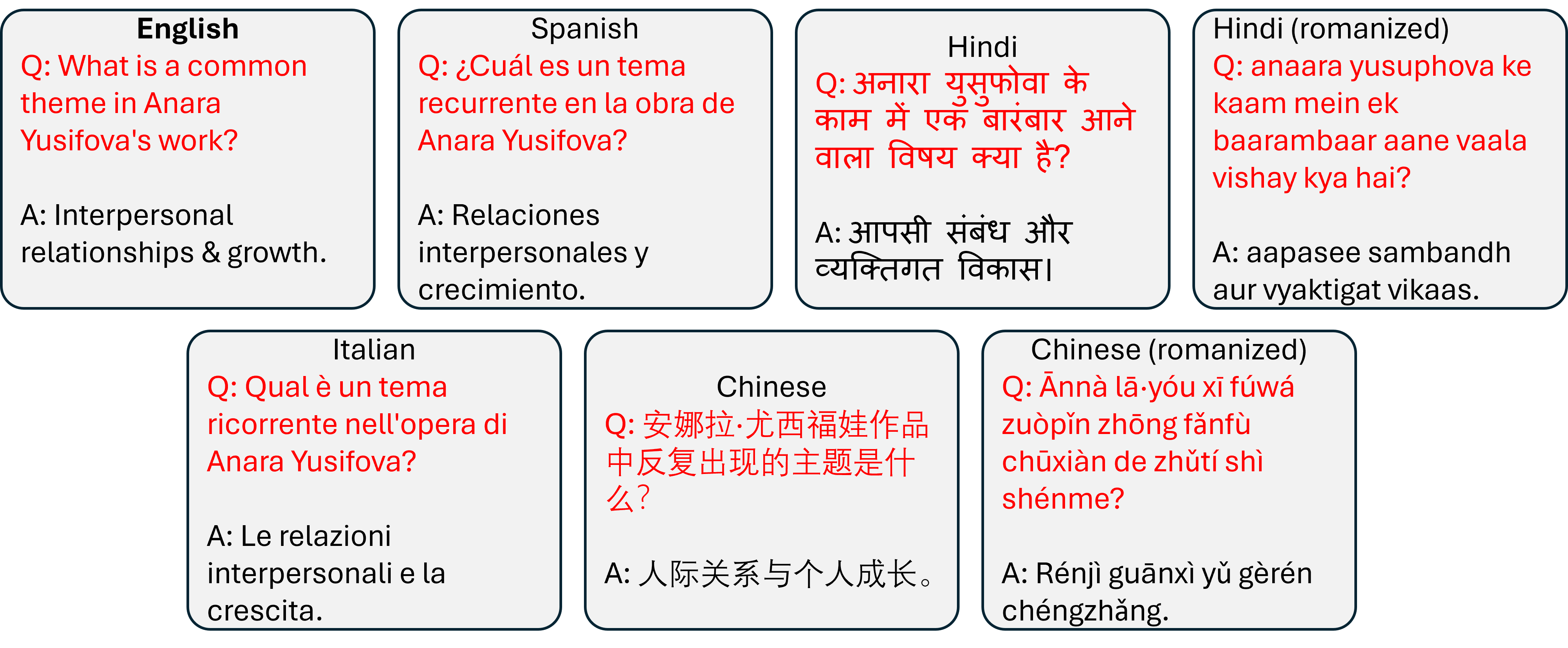}
  \end{overpic}
  \caption{Examples of question-answer pairs from all seven translations of the TOFU dataset.}
    \label{fig:translations}
\end{figure}

To study cross-lingual unlearning, we construct multilingual versions of the TOFU benchmark \citep{maini2024tofutaskfictitiousunlearning}. TOFU is an English-only dataset composed of synthetic factual statements about fictional authors, designed to probe memorization by pairing each author with a single, templated fact that must be removed. It also includes a retain set concerning real authors and world knowledge, allowing controlled measurement of collateral damage. We generate parallel translations in five languages: English, Chinese, Hindi, Italian, and Spanish, along with romanized versions of Hindi and Chinese (see Figure \ref{fig:translations}). Each translated instance preserves the original fact structure and answer template, enabling consistent evaluation across languages. These experiments use the 10\% forget set size due to the greater difficulty with successful unlearning (see Appendix \ref{forget_set_size} for details).

\paragraph{Models.}
We evaluate three families of pretrained language models with varying degrees of multilingual capability.  
Mistral, Mixtral, and Magistral form a progression from an English-only model to partially multilingual support (Italian, Spanish, and Chinese) to full support across English, Chinese, Italian, and Spanish.  
Similarly, the Llama family includes Llama 2 (English-only) and Llama 3 and Llama 4 (multilingual models covering English, Hindi, Italian, and Spanish). Finally, we include Aya-23, a natively multilingual model explicitly trained for broad cross-lingual coverage and generalization, providing a complementary contrast to the incremental multilingual extensions represented by the Mistral and Llama families.
Together, these model families enable controlled comparison across monolingual, partially multilingual, and fully multilingual architectures with differing training philosophies.
\begin{figure*}[t]
    \centering
  \begin{overpic}[width=0.8\linewidth]{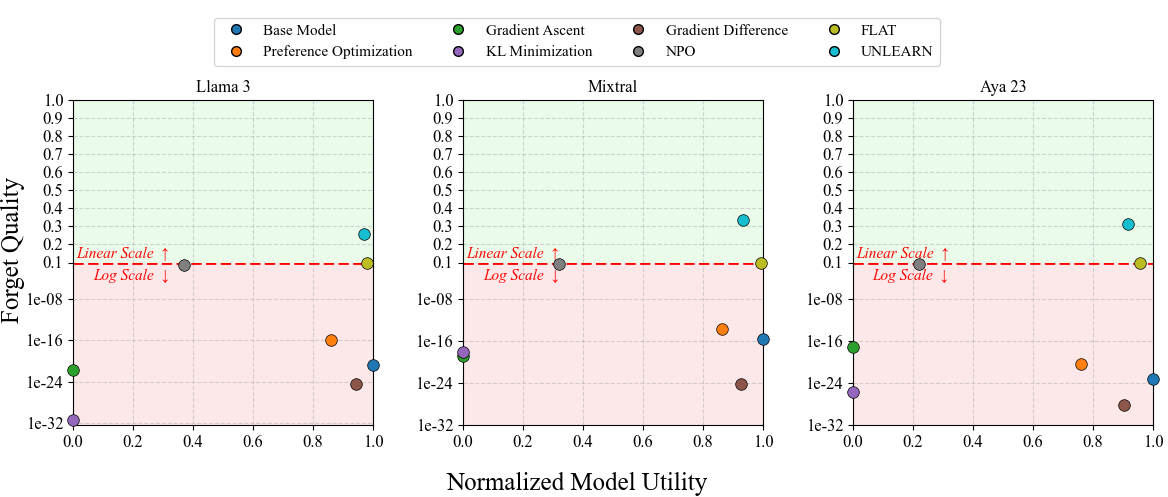}
  \end{overpic}
  \caption{Comparison of unlearning performance on the English-English TOFU benchmark \cite{maini2024tofutaskfictitiousunlearning} for three of our models: Llama 3, Mixtral, and Aya 23. See Table \ref{tab:english_english} for full results.}
    \label{fig:tofu_results}
\end{figure*}
\paragraph{Unlearning Methods.}
We evaluate several representative approaches from the major families of unlearning methods:  
gradient ascent \citep{jang2023knowledge}, KL-based retain-objective unlearning \citep{wang2025balancing}, gradient-difference corrections \citep{liu2022continual}, Negative Preference Optimization (NPO) \citep{zhang2024npo}, FLAT \citep{wang2024flat}, and a subspace-projection method \textsc{UNLEARN} \citep{lizzo2025unlearn, lizzo2025patent}.

\paragraph{Metrics.}
We adopt the two evaluation metrics defined in TOFU.  
Forget quality is measured as a two-sample p-value comparing the unlearned model's outputs on the forget set to the original model's outputs, where values above $0.1$ indicate successful forgetting under the TOFU criterion.  
Model utility is computed as the harmonic mean of nine capability metrics and is reported relative to the base model so that a value of 1.0 corresponds to unchanged performance.

\subsection{Experimental Configurations}

We evaluate cross-lingual unlearning with a set of standardized configurations that are reused across multiple experimental analyses through the paper. These configurations broadly define how forgetting is applied across languages, independent of the specific model family, baseline, or analysis focus. These core configurations are One-to-One Unlearning, Many-to-One Unlearning, and Transliteration, each of which isolate a different aspect of multilingual forgetting.

    \noindent {\bf One-to-One Unlearning:} unlearning is performed in a single source language, and the resulting model is evaluated on all other languages supported by that model family. Four analyses center on this configuration: a base exploration of English-to-English unlearning, a look at the failure of cross-lingual unlearning for \textbf{monolingual} models, the performance of cross-lingual unlearning for \textbf{multilingual} models, and a look at how model unlearning performance evolves with newer models.
    
    \noindent {\bf Many-to-One Unlearning:} unlearning is performed jointly over multiple languages and evaluated on a held-out target language. This design tests whether multilingual supervision during unlearning encourages more language-agnostic removal of the target facts and whether additional languages help disambiguate shared representational structure.
    
    \noindent {\bf Transliteration:} for the two languages with non-Latin scripts (Hindi and Chinese), we test how orthographic form influences unlearning. We create transliterated versions of the Chinese and Hindi TOFU datasets using standard romanization schemes. Unlearning is performed separately on both the original-script datasets and on the transliterated versions, and each model is evaluated in both forms. This setting probes whether script-level tokenization differences affect how knowledge is stored and whether forgetting transfers between these variants of the same language.

Together, these three configurations allow us to evaluate unlearning under complementary cross-lingual conditions and to probe how multilingual structure and tokenization influence the propagation of forgetting.
\begin{figure}
    \centering
  \begin{overpic}[width=0.98\linewidth]{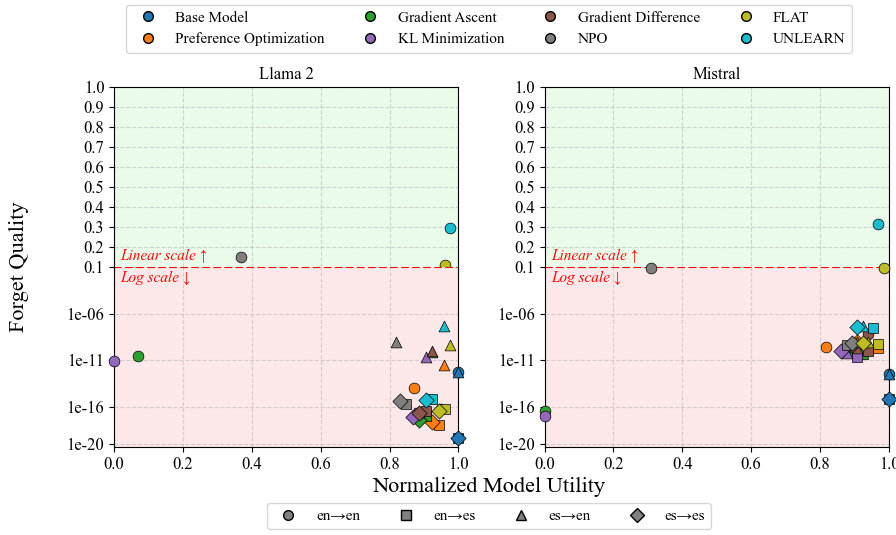}
  \end{overpic}
  \caption{Unlearning performance for non-multilingual models (Llama-2, Mistral) evaluated on English/Spanish TOFU dataset for all unlearning algorithms. See Table \ref{tab:crosslingual_gen1} for full results.}
    \label{fig:llama2_mistral}
\end{figure}
\subsection{Results and Analysis}
\begin{figure*}[t]
    \centering
  \includegraphics[width=12cm,height=5cm]{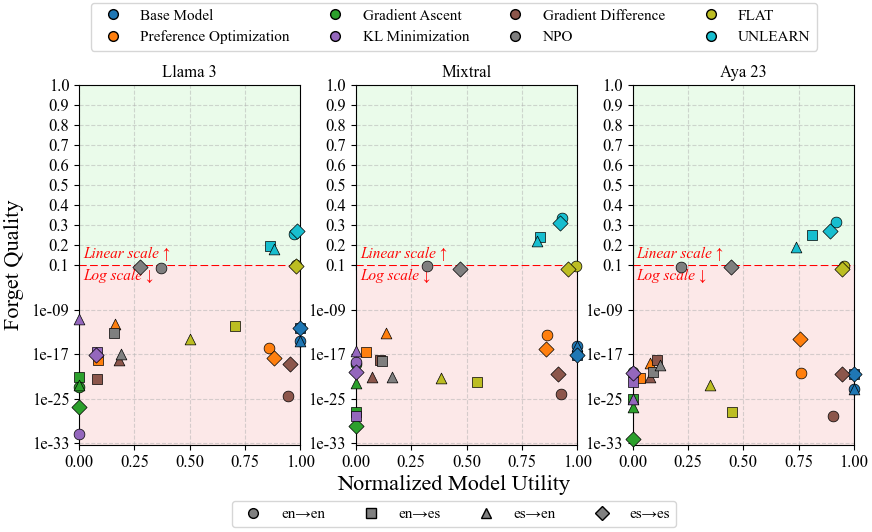}
  \caption{Unlearning  for multilingual models (Llama-3, Mixtral, Aya-23) evaluated on English/Spanish TOFU. See Table \ref{tab:crosslingual_gen2} for full results.}
    \label{fig:llama3_mixtral}
\end{figure*}
We now present the results of the three experimental configurations described above. Each subsection summarizes the quantitative outcomes, followed by analysis of cross-lingual behavior and model-specific trends. All evaluations use the TOFU forget-quality and model-utility metrics.

\paragraph{English-to-English Unlearning.}
We begin by evaluating unlearning in the simplest setting: forgetting and evaluating in English using Llama 3, Mixtral, and Aya-23. Figure \ref{fig:tofu_results} shows that all classical methods (gradient ascent, KL-based objectives, gradient-difference variants, preference optimization, etc.) fail to achieve statistically significant forgetting. FLAT approaches the statistical threshold but does not consistently cross it. NPO attains strong forget quality but substantially degrades utility as the size of the forget set increases. UNLEARN achieves statistically significant forgetting while maintaining utility near 1.0, illustrating a favorable trade-off that will serve as a reference point for the multilingual experiments moving forward.

\paragraph{Cross-Lingual Baselines in Non-Multilingual Models.}
Next, we examine cross-lingual forgetting in our two base models that were not designed for multilingual use: Mistral and Llama 2. Figure \ref{fig:llama2_mistral} shows English $\leftrightarrow$ Spanish on both of these models. Because these models lack meaningful capability in Spanish, there are two emergent trends: 1) Forget quality is poor regardless of model, direction of learning, or unlearning algorithm and 2) Model utility remains high regardless of setting.

These two trends, when taken together, reveal that the finetuned models overfit to the information learned in unknown languages, a trend that persists across all language pairs for these models. As such, analysis on monolingual LLMs is flawed; deeper analysis focuses on the newer models that feature explicit multilingual training.

\begin{figure*}
    \centering
  \includegraphics[width=12cm,height=6cm]{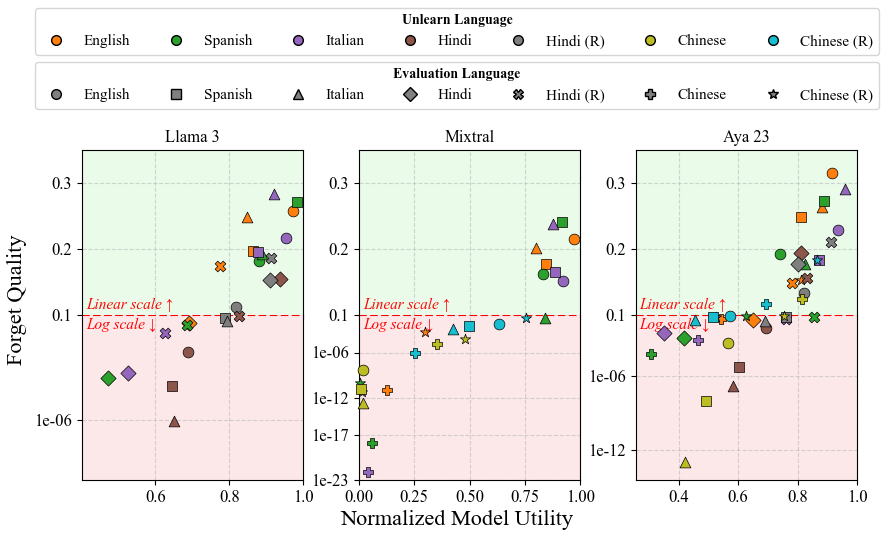}
  \caption{UNLEARN for multilingual models (Llama-3, Mixtral, Aya-23) on all languages in TOFU. See Table \ref{tab:crosslingual_all} for full results.}
    \label{fig:onetoone_unlearn}
\end{figure*}

\paragraph{Cross-Lingual Forgetting in Multilingual Models}
We turn our analysis to models that are designed for multilinguality: Mixtral,  Llama 3, and Aya-23. Unlike the baselines above, these models possess meaningful skill in multiple languages. Figure \ref{fig:llama3_mixtral} examines symmetric English $\leftrightarrow$ Spanish across all algorithms, while Figure \ref{fig:onetoone_unlearn} shows the subspace-projection algorithm UNLEARN applied to all language pairs.

Across all model families, three patterns consistently appear:  1) Most methods completely degrade performance across any language pairs. UNLEARN is the only algorithm that still attains statistically sigificant forgetting across any language pair, hence the focus on this algorithm. 2) There is worse model utility and forget quality with the two non-latin script languages. Chinese and Hindi show lower forget quality and model utility. 3) Cross-lingual unlearning is assymetric. Forgetting facts in English and evaluating in other languages tends to have higher forget quality than the reverse. This is likely explained by the higher representation of English in the training sets for both models.

\begin{comment}
    
\begin{figure}
    \centering
  \begin{overpic}[width=0.98\linewidth]{img/english_spanish.png}
  \end{overpic}
  \caption{Unlearning performance for multilingual models (Llama-3, Mixtral, Aya-23) evaluated on English/Spanish TOFU data for all unlearning algorithms.}
    \label{fig:llama3_mixtral}
\end{figure}

\begin{figure}
    \centering
  \begin{overpic}[width=0.98\linewidth]{img/multilingual_unlearn.png}
  \end{overpic}
  \caption{Unlearning performance for multilingual models (Llama-3, Mixtral, Aya-23) evaluated on all languages in TOFU dataset, using UNLEARN algorithm.}
    \label{fig:onetoone_unlearn}
\end{figure}
\end{comment}

\begin{figure}
    \centering
  \begin{overpic}[width=0.98\linewidth]{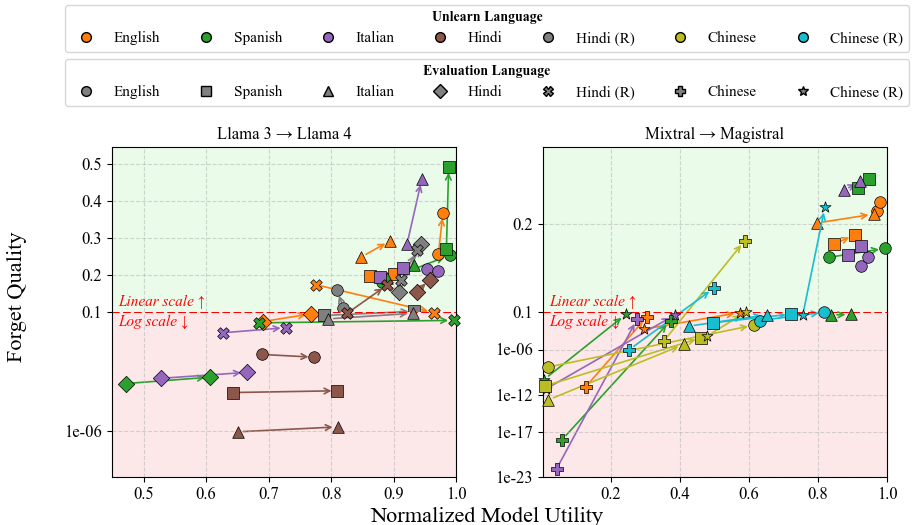}
  \end{overpic}
  \caption{Unlearning across model generations. See Tables \ref{tab:generations_llama}, \ref{tab:generations_mixtral} for full results.}
  \label{fig:generations}
\end{figure}

\paragraph{Effect of Model Generation on Cross-Lingual Unlearning}
Figure \ref{fig:generations} compares successive generations of multilingual models: Mixtral $\rightarrow$ Magistral and Llama 3 $\rightarrow$ Llama 4 on identical settings. There is an obvious boost in cross-lingual unlearning performance across unlearning methods and languages pairs. Normalized model utility increases almost universally across the board, and several forget-quality scores rise above the statistically significant ($\alpha=0.1$) threshold. Because both metrics are computed relative to the corresponding base model, these gains reflect genuine improvements in relative cross-lingual forgetting ability. This suggests that the feasibility of cross-lingual unlearning depends not only on the unlearning algorithm but also on the underlying multilingual strength of the model itself. 

\begin{figure*}
    \centering
 \includegraphics[width=12cm,height=6cm]{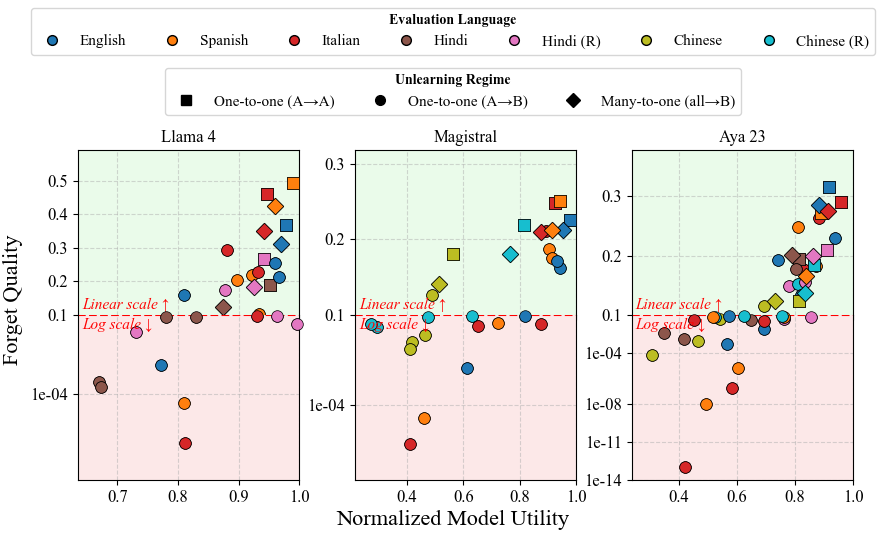}
  \caption{Unlearning performance for multilingual models comparing one-to-one vs many-to-one unlearning. See Tables \ref{tab:manytomany_llama4}, \ref{tab:manytomany_magistral}, \ref{tab:manytomany_aya23} for full results.}
    \label{fig:manytoone}
\end{figure*}

\paragraph{Many-to-One Unlearning}
In the many-to-one setting, unlearning is performed jointly across all languages except one held-out-language. In Figure \ref{fig:manytoone}, we observe a consistent improvement over the one-to-one results, a trend we notice for \textbf{every single} held-out language. Additionally, there is a notable clustering of points, indicating that multilingual unlearning produces a more stable and language-agnostic forgetting signal. As we will discuss and look into later, these behaviors point to the existence of an interlingua space within the weight space of the LLMs.

\begin{figure}
    \centering
 \includegraphics[width=8cm,height=4cm]{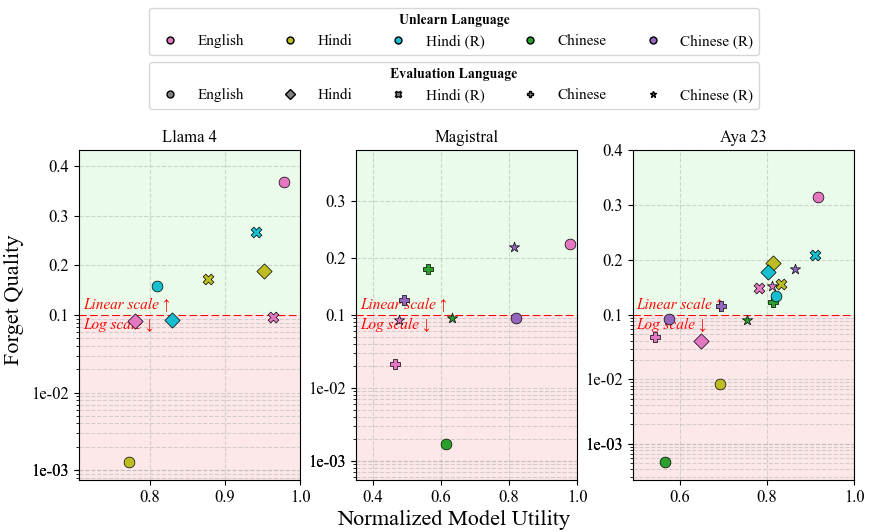}
  \caption{Unlearning: languages with different scripts. See Table \ref{tab:transliteration} for full results.}
    \label{fig:transliteration}
\end{figure}

\paragraph{Transliteration Effects.}
We next evaluate cross-lingual unlearning under script changes for both Hindi and Chinese. A clear pattern emerges in Figure \ref{fig:transliteration}: English~$\rightarrow$~romanized Hindi yields stronger forgetting than English~$\rightarrow$~Devanagari, and romanized Hindi~$\rightarrow$~Devanagari also outperforms English~$\rightarrow$~Devanagari. Similar trends hold for Chinese and its pinyin transliteration. These results suggest that transliteration reduces orthographic distance and improves transfer during unlearning, while the remaining performance gap indicates that some components of factual representation remain tied to script-specific subspaces. Together, these findings reinforce the presence of an interlingual semantic structure while showing that script continues to shape how multilingual models store and forget knowledge.

\section{Interlingua Discussion}
\label{discussion}

\subsection{Evidence for a Multilingual Interlingua}
Across all experimental configurations, the results consistently indicate that multilingual models encode a shared representational structure that spans multiple languages. Three empirical observations support this conclusion.

First, in the one-to-one unlearning setting, forgetting often transfers, at least in part, from the source language to others. Although the transfer is not uniform, the presence of any forgetting signal in languages that were not explicitly targeted implies that factual knowledge is stored in shared semantic regions rather than in language-isolated subspaces.

Second, the many-to-one experiments show that jointly unlearning multiple languages yields both higher forget quality and a more compact distribution of results across held-out languages. This pattern suggests that multilingual supervision during unlearning strengthens access to a language-agnostic component of the representation, producing a more stable and consistently applied forgetting operation.

Third, the transliteration experiments demonstrate that script changes influence cross-lingual forgetting but do not eliminate it. Romanized Hindi and pinyin-transliterated Chinese display stronger transfer than their original-script forms, yet both variants still participate in cross-lingual forgetting. This indicates that factual representations reside primarily in a shared semantic space, while script-specific tokenization contributes additional language-dependent variation.

Taken together, these findings support the existence of a multilingual interlingua: a shared, low-dimensional subspace that aligns meaning across languages and through which forgetting propagates.

\subsection{Visualizing the Interlingua via Subspace Geometry}
To characterize the multilingual structure underlying cross-lingual unlearning, we analyze the task-specific subspaces extracted by the subspace-projection method \textsc{UNLEARN} for each language. The discrimination procedure isolates the directions in weight space most responsible for encoding the targeted factual association by contrasting task-induced updates with updates from unrelated control tasks, yielding a low-rank basis for each language.

\begin{figure}
    \centering
\includegraphics[width=4cm,height=4cm]{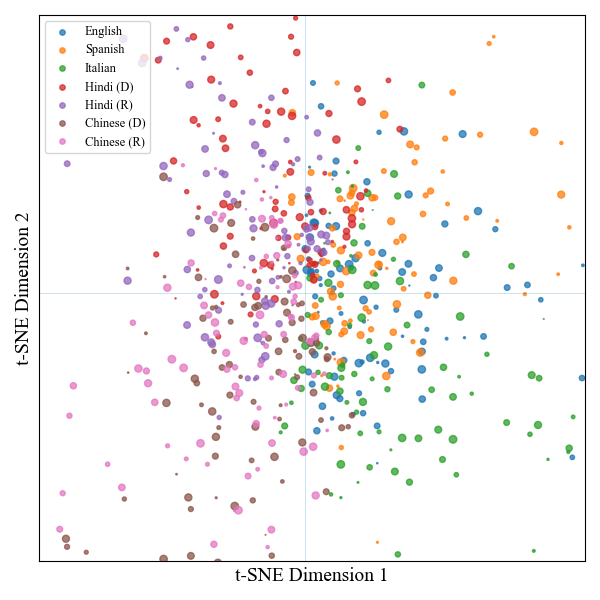}
    \caption{
    t-SNE of task-specific subspace components extracted by subspace-projection in Aya-23 across languages. Each point represents a basis vector from a language-specific subspace, colored by supervision language; marker size reflects singular-value rank (larger = stronger contribution).
 %   The substantial overlap across languages indicates the presence of a shared interlingual subspace.
    }
    \label{fig:tsne}
\end{figure}

Figure~\ref{fig:tsne} presents a t-SNE visualization of these subspaces, revealing substantial overlap across languages. This structure reveals a natural decomposition of the task representation into (i) a shared semantic subspace that is stable across languages and encodes language-independent factual content, and (ii) a language-specific residual subspace reflecting orthographic, tokenization, and morphology-driven variation. The dominance of shared directions provides a direct geometric explanation for the cross-lingual transfer of forgetting observed earlier.

\subsection{Intervention Experiments: Causal Evidence for the Interlingua}
\begin{figure*}
    \centering
\includegraphics[width=12cm,height=5cm]{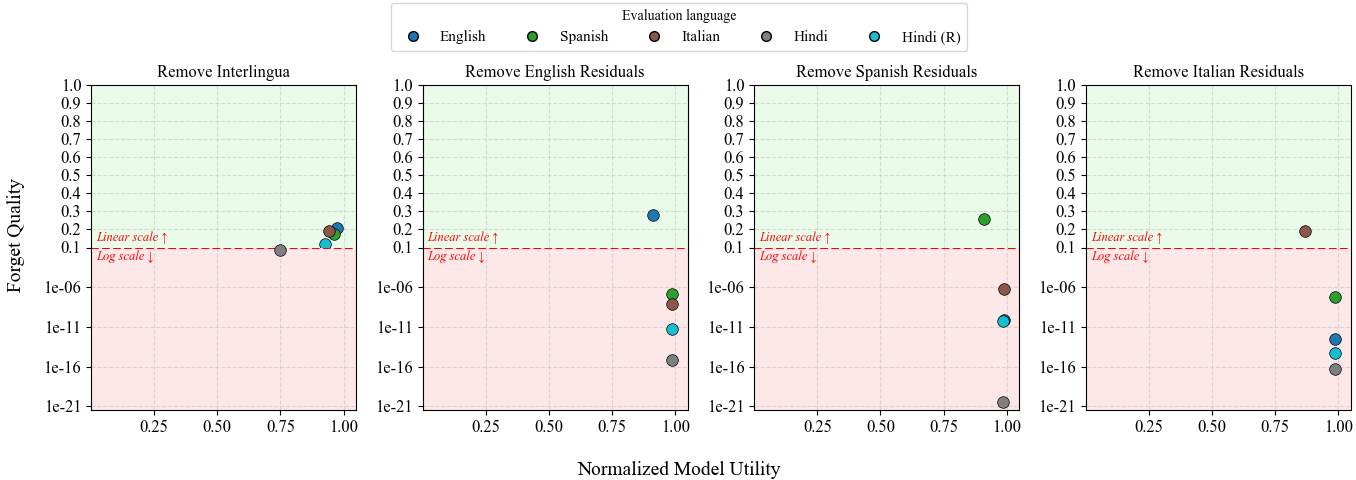}
  \caption{A comparison showing removal of the interlingua space vs the language residual subspaces.}
    \label{fig:interlingua_removal}
\end{figure*}

While subspace similarity measurements provide correlational evidence, we further perform causal ablations to validate the proposed decomposition into shared and language-specific components. Using the discriminated bases returned by \textsc{UNLEARN}, we construct two targeted interventions whose effects are shown in Figure~\ref{fig:interlingua_removal}.\\

\noindent{\bf Removing the Shared Interlingua Subspace}
We first identify the shared interlingua of all languages by extracting the overlapping  weight spaces using \textsc{UNLEARN}'s discrimination process. Projecting out this shared interlingua produces a nearly uniform effect across languages: all languages exhibit statistically significant forget quality and retain strong model utility. In other words, removing the shared subspace causes the fact to be forgotten across all languages. This behavior indicates that this shared component is the primary carrier of language-agnostic factual structure.\\

\noindent{\bf Removing Language-Specific Residual Subspaces}
 Next, for each language, we isolate the residual subspace, the weights for that language orthogonal to the shared interlingua. Removing these residuals has a sharply different effect. Only the targeted language shows high forget quality, accompanied by a modest reduction in model utility. All other languages have nearly perfect normalized model utility and little to no forgetting signal. This selective collapse confirms that each language's residual subspace contributes a small, language-dependent component that is still import for factual representation.

Together, these two interventions provide direct causal evidence for the interlingua hypothesis. The multilingual encoding of factual knowledge consists of a shared semantic subspace that governs cross-lingual transfer, augmented by smaller language-specific residuals tied to script, tokenization, and morphology. This decomposition both explains the patterns observed throughout our experiments and clarifies why subspace-projection approaches are well-suited for multilingual unlearning.

\section{Conclusions}
\label{conclusions}

This work presents the first systematic study of cross-lingual unlearning in large language models. Using multilingual extensions of the TOFU benchmark, we show that most existing unlearning methods fail to remove targeted knowledge beyond the language in which unlearning is performed, even in strongly multilingual models. These results reveal challenges in cross-lingual forgetting absent in monolingual settings.

In contrast, subspace-projection (UNLEARN) achieves reliable cross-lingual forgetting while preserving overall model utility. Across languages and model families, it propagates forgetting into held-out languages without the collateral degradation seen in alternative approaches.

Our analysis explains why. Knowledge in multilingual LLMs exists across a shared, language-independent semantic subspace alongside residual components driven by script, tokenization, and morphology. This shared interlingua explains observed cross-lingual patterns, including partial transfer under transliteration.

Causal ablations confirm this structure: removing the shared semantic subspace produces strong forgetting across all languages while maintaining utility, whereas removing a language-specific residual yields targeted forgetting confined to that language. These results show that multilingual representations combine a common semantic core with language-dependent variations.

Overall, our findings indicate that effective cross-lingual unlearning requires operating on representational structure rather than relying solely on loss-based fine-tuning, underscoring the need for multilingual evaluation. As safety, privacy, and regulatory demands increasingly require targeted knowledge removal, understanding multilingual representation is essential for controllable and responsible systems.

\section*{Limitations}
While our experiments span multiple languages and scripts, the evaluated set remains limited compared to the full linguistic diversity seen in real-world multilingual systems. In particular, low-resource languages, highly agglutinative or polysynthetic languages, and writing systems not represented here may exhibit different unlearning behavior. Further, this work also focuses on factual knowledge removal as defined by the TOFU benchmark and its multilingual extensions, and other aspects of model behavior, such as procedural knowledge, conversational norms, or higher-level reasoning, may be encoded differently and may not follow the same subspace structure. 

\section*{Ethics Statement}
This work investigates methods for targeted knowledge removal in multilingual large language models, motivated by safety, privacy, and regulatory considerations. Our experiments rely on existing benchmarks and pretrained models and do not involve human subjects or direct interaction with users. The proposed unlearning techniques are intended to support responsible model deployment by enabling more precise control over model behavior, such as removing outdated, incorrect, or sensitive information.

At the same time, methods for modifying model representations could be misused to suppress legitimate information or introduce bias if applied without appropriate oversight. We emphasize that unlearning should be guided by clear governance and transparency, and that technical mechanisms alone are not sufficient to address broader ethical concerns. We also report results across multiple languages to highlight differences in unlearning behavior and to avoid assumptions that safety interventions transfer uniformly across linguistic contexts.

\section*{AI Statement}
We used AI-based tools to assist with writing and editing, such as improving clarity and organization of the text. All ideas, experiments, results, and conclusions are the authors’ own, and all content was reviewed and finalized by the authors. AI tools were not used to generate data or make scientific decisions.

\bibliography{emnlp2023}
\bibliographystyle{acl_natbib}

\appendix

\begin{figure}[b]
  \centering
  \begin{overpic}[width=0.9\linewidth]{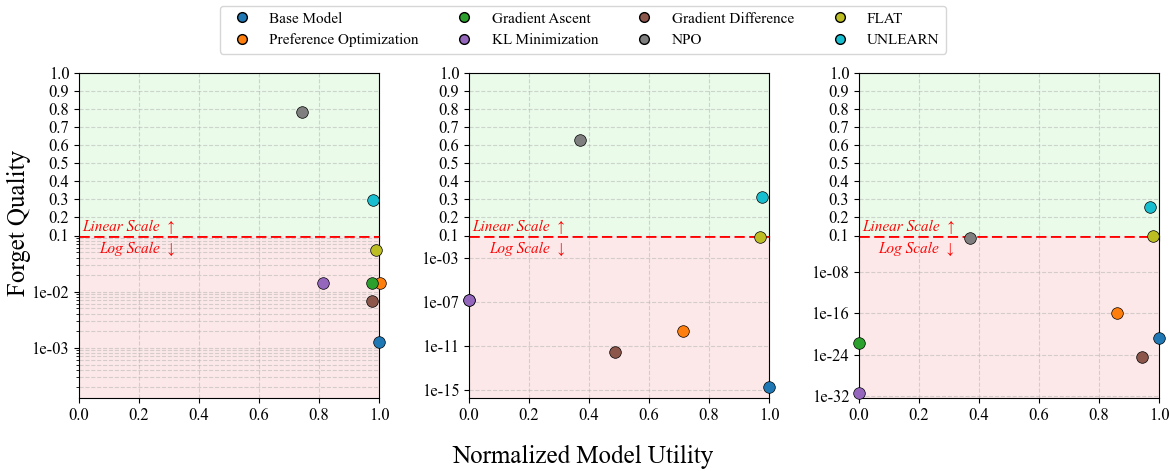}
  \end{overpic}
  \caption{Comparison of unlearning performance on the TOFU benchmark \cite{maini2024tofutaskfictitiousunlearning} across forget-set sizes (1\%, 5\%, 10\%). Each point represents a model's trade-off between normalized model utility (higher is better) and forget quality (p-value; higher is better).  The dashed red line marks the $\alpha = 0.1$ threshold distinguishing statistically significant forgetting (above the line) from unsuccessful forgetting (below the line), as well as a transition between linear and log scales for forget quality.}
    \label{fig:tofu_forget_size}
\end{figure}
\section{TOFU Forget Set Size Selection }
\label{forget_set_size}
In our main analysis, we report results using a forget set size of 10\%. This choice was made because larger forget sets consistently represent the most challenging unlearning regime across models and methods. Empirically, we observe that unlearning difficulty increases monotonically with forget set size, with 10\% producing the largest degradation in forget quality and the strongest tension with model utility.
This trend is illustrated in Figure \ref{fig:tofu_forget_size}, which shows forget quality as a function of forget set size, and is further corroborated by the full results table in Table \ref{tab:forget_size}. Since conclusions drawn at 10\% reflect behavior at smaller forget sizes, focusing on this setting provides a conservative and informative evaluation of unlearning performance.

\section{Full Results}
The rest of the appendix contains full tables for the results obtained throughout the paper.
\label{full_results_table}
\begin{table*}[!t]
\centering
\small
\begin{tabular}{lcccccc}
\toprule
& \multicolumn{2}{c}{\textbf{Llama 3}} 
& \multicolumn{2}{c}{\textbf{Mixtral}} 
& \multicolumn{2}{c}{\textbf{Aya-23}} \\
\cmidrule(lr){2-3}\cmidrule(lr){4-5}\cmidrule(lr){6-7}\\
\textbf{Method} 
& Utility $\uparrow$ & Forget $\uparrow$ 
& Utility $\uparrow$ & Forget $\uparrow$ 
& Utility $\uparrow$ & Forget $\uparrow$ \\
\midrule
Base Model                & 0.6227 & $1.83{\times}10^{-21}$ & 0.5912 & $3.0{\times}10^{-16}$ & 0.5503 & $6.4{\times}10^{-24}$ \\
\midrule
Preference Optimization  & 0.5357 & $1.06{\times}10^{-16}$ & 0.5100 & $2.5{\times}10^{-14}$ & 0.4182 & $4.3{\times}10^{-21}$ \\
Gradient Ascent           & 0.0000 & $1.43{\times}10^{-22}$ & 0.0000 & $1.35{\times}10^{-19}$ & 0.0000 & $6.28{\times}10^{-18}$ \\
KL Minimization           & 0.0000 & $4.07{\times}10^{-32}$ & 0.0000 & $7.23{\times}10^{-19}$ & 0.0000 & $2.38{\times}10^{-26}$ \\
Gradient Difference       & 0.5872 & $2.83{\times}10^{-25}$ & 0.5483 & $5.83{\times}10^{-25}$ & 0.4978 & $6.19{\times}10^{-29}$ \\
NPO                       & 0.2297 & 0.0291                & 0.1893 & 0.0587                & 0.1213 & 0.0468 \\
FLAT                      & \textbf{0.6104} & 0.0841                & \textbf{0.5871} & 0.0692                & \textbf{0.5264} & 0.0765 \\
UNLEARN                   & 0.6047 & \textbf{0.257} \checkmark        & 0.5513 & \textbf{0.334} \checkmark        & 0.5047 & \textbf{0.315} \checkmark \\
\bottomrule
\end{tabular}
\caption{Comparison of unlearning performance on the English-English TOFU benchmark \cite{maini2024tofutaskfictitiousunlearning} for three models: Llama 3, Mixtral, and Aya 23.}
\label{tab:english_english}
\end{table*}

\begin{table*}[t]
\centering
\small
\setlength{\tabcolsep}{5pt}
\begin{tabular}{|l|cc|cc|cc|cc|}
\hline
\textbf{Method}
& \multicolumn{2}{c|}{\textbf{en$\to$en}}
& \multicolumn{2}{c|}{\textbf{en$\to$es}}
& \multicolumn{2}{c|}{\textbf{es$\to$en}}
& \multicolumn{2}{c|}{\textbf{es$\to$es}} \\
& Utility $\uparrow$ & Forget $\uparrow$
& Utility $\uparrow$ & Forget $\uparrow$
& Utility $\uparrow$ & Forget $\uparrow$
& Utility $\uparrow$ & Forget $\uparrow$ \\
\hline
\multicolumn{9}{|l|}{\textbf{Llama 2}} \\
\hline
Base Model
& 0.5743 & $4.78{\times}10^{-13}$
& 0.0536 & $4.43{\times}10^{-20}$
& 0.5743 & $4.78{\times}10^{-13}$
& 0.0532 & $4.43{\times}10^{-20}$ \\
\hline
Preference Optimization
& 0.4996 & $1.34{\times}10^{-14}$
& 0.0507 & $1.62{\times}10^{-18}$
& 0.5539 & $3.47{\times}10^{-12}$
& 0.0494 & $2.58{\times}10^{-18}$ \\
Gradient Ascent
& 0.0400 & $3.81{\times}10^{-11}$
& 0.0485 & $9.26{\times}10^{-18}$
& 0.5364 & $7.92{\times}10^{-11}$
& 0.0478 & $4.63{\times}10^{-18}$ \\
KL Minimization
& 0.0000 & $8.41{\times}10^{-12}$
& 0.0476 & $2.93{\times}10^{-17}$
& 0.5287 & $2.68{\times}10^{-11}$
& 0.0469 & $7.51{\times}10^{-18}$ \\
Gradient Difference
& 0.5456 & $6.83{\times}10^{-17}$
& 0.0483 & $3.57{\times}10^{-17}$
& 0.5341 & $1.46{\times}10^{-10}$
& 0.0474 & $2.81{\times}10^{-17}$ \\
NPO
& 0.2125 & \textbf{0.153}$^{\checkmark}$
& 0.0459 & $2.47{\times}10^{-16}$
& 0.4728 & $9.84{\times}10^{-10}$
& 0.0443 & $4.66{\times}10^{-16}$ \\
FLAT
& 0.5513 & \textbf{0.113}$^{\checkmark}$
& \textbf{0.0516} & $5.92{\times}10^{-17}$
& \textbf{0.5634} & $4.51{\times}10^{-10}$
& \textbf{0.0508} & $3.79{\times}10^{-17}$ \\
UNLEARN
& \textbf{0.5687} & \textbf{0.297}$^{\checkmark}$
& 0.0495 & $7.36{\times}10^{-16}$
& 0.5569 & $5.84{\times}10^{-8}$
& 0.0486 & $5.41{\times}10^{-16}$ \\
\hline
\multicolumn{9}{|l|}{\textbf{Mistral}} \\
\hline
Base Model
& 0.6579 & $3.42{\times}10^{-13}$
& 0.6548 & $6.71{\times}10^{-16}$
& 0.6526 & $3.42{\times}10^{-13}$
& 0.6583 & $6.71{\times}10^{-16}$ \\
\hline
Preference Optimization
& 0.5312 & $2.58{\times}10^{-10}$
& 0.6342 & $2.83{\times}10^{-10}$
& 0.6076 & $6.91{\times}10^{-10}$
& 0.5968 & $8.54{\times}10^{-10}$ \\
Gradient Ascent
& 0.0000 & $3.71{\times}10^{-17}$
& 0.6039 & $5.64{\times}10^{-11}$
& 0.5821 & $1.84{\times}10^{-10}$
& 0.5784 & $2.53{\times}10^{-10}$ \\
KL Minimization
& 0.0000 & $9.28{\times}10^{-18}$
& 0.5947 & $2.71{\times}10^{-11}$
& 0.5719 & $6.44{\times}10^{-11}$
& 0.5693 & $1.62{\times}10^{-10}$ \\
Gradient Difference
& 0.6164 & $7.92{\times}10^{-9}$
& 0.6117 & $1.46{\times}10^{-10}$
& 0.5984 & $2.87{\times}10^{-10}$
& 0.5829 & $4.68{\times}10^{-10}$ \\
NPO
& 0.2015 & 0.084
& 0.5726 & $4.38{\times}10^{-10}$
& 0.5949 & $9.41{\times}10^{-10}$
& 0.5896 & $6.72{\times}10^{-10}$ \\
FLAT
& \textbf{0.6468} & 0.072
& \textbf{0.6387} & $5.86{\times}10^{-10}$
& \textbf{0.6134} & $8.29{\times}10^{-10}$
& \textbf{0.6038} & $7.91{\times}10^{-10}$ \\
UNLEARN
& 0.6361 & \textbf{0.315}$^{\checkmark}$
& 0.6289 & $3.64{\times}10^{-8}$
& 0.6031 & $5.73{\times}10^{-8}$
& 0.5969 & $4.38{\times}10^{-8}$ \\
\hline
\end{tabular}
\caption{Cross-lingual unlearning results for the first generation of models. Best utility per column for a given model is bolded. Forget quality values exceeding $\alpha=0.1$ are bolded and marked with a checkmark. These models are monolingual, so it is no surprise that every unlearning method struggles.}
\label{tab:crosslingual_gen1}
\end{table*}

\begin{table*}[t]
\centering
\small
\setlength{\tabcolsep}{5pt}
\begin{tabular}{|l|cc|cc|cc|cc|}
\hline
\textbf{Method}
& \multicolumn{2}{c|}{\textbf{en$\to$en}}
& \multicolumn{2}{c|}{\textbf{en$\to$es}}
& \multicolumn{2}{c|}{\textbf{es$\to$en}}
& \multicolumn{2}{c|}{\textbf{es$\to$es}} \\
& Utility $\uparrow$ & Forget $\uparrow$
& Utility $\uparrow$ & Forget $\uparrow$
& Utility $\uparrow$ & Forget $\uparrow$
& Utility $\uparrow$ & Forget $\uparrow$ \\
\hline

\multicolumn{9}{|l|}{\textbf{Llama 3}} \\ \hline
Base Model
& 0.6227 & $2.37{\times}10^{-15}$
& 0.5919 & $4.31{\times}10^{-13}$
& 0.6227 & $2.37{\times}10^{-15}$
& 0.5919 & $4.31{\times}10^{-13}$ \\
\hline
Preference Optimization
& 0.5357 & $1.06{\times}10^{-16}$
& 0.0506 & $1.42{\times}10^{-18}$
& 0.1025 & $3.58{\times}10^{-12}$
& 0.5216 & $2.84{\times}10^{-18}$ \\
Gradient Ascent
& 0.0000 & $1.35{\times}10^{-23}$
& 0.0000 & $7.74{\times}10^{-22}$
& 0.0000 & $2.98{\times}10^{-23}$
& 0.0000 & $3.25{\times}10^{-27}$ \\
KL Minimization
& 0.0000 & $4.07{\times}10^{-32}$
& 0.0478 & $2.63{\times}10^{-17}$
& 0.0000 & $2.91{\times}10^{-11}$
& 0.0467 & $7.84{\times}10^{-18}$ \\
Gradient Difference
& 0.5872 & $2.83{\times}10^{-25}$
& 0.0484 & $3.74{\times}10^{-22}$
& 0.1136 & $1.47{\times}10^{-18}$
& 0.5638 & $2.96{\times}10^{-19}$ \\
NPO
& 0.2297 & 0.0291
& 0.0941 & $5.63{\times}10^{-14}$
& 0.1173 & $9.46{\times}10^{-18}$
& 0.1639 & 0.0517 \\
FLAT
& \textbf{0.6104} & 0.0841
& 0.4174 & $1.73{\times}10^{-12}$
& 0.3118 & $4.52{\times}10^{-15}$
& 0.5802 & 0.0712 \\
UNLEARN
& 0.6047 & \textbf{0.257}$^{\checkmark}$
& \textbf{0.5105} & \textbf{0.197}$^{\checkmark}$
& \textbf{0.5482} & \textbf{0.182}$^{\checkmark}$
& \textbf{0.5824} & \textbf{0.271}$^{\checkmark}$ \\
\hline

\multicolumn{9}{|l|}{\textbf{Mixtral}} \\ \hline
Base Model
& 0.5912 & $3.48{\times}10^{-16}$
& 0.5393 & $6.91{\times}10^{-18}$
& 0.5912 & $3.48{\times}10^{-16}$
& 0.5393 & $6.91{\times}10^{-18}$ \\
Preference Optimization
& 0.5106 & $2.53{\times}10^{-14}$
& 0.0246 & $2.71{\times}10^{-17}$
& 0.0815 & $6.48{\times}10^{-14}$
& 0.4621 & $8.47{\times}10^{-17}$ \\
Gradient Ascent
& 0.0000 & $1.35{\times}10^{-19}$
& 0.0000 & $4.74{\times}10^{-28}$
& 0.0000 & $7.73{\times}10^{-23}$
& 0.0000 & $1.25{\times}10^{-30}$ \\
KL Minimization
& 0.0000 & $4.07{\times}10^{-19}$
& 0.0000 & $8.00{\times}10^{-29}$
& 0.0000 & $3.61{\times}10^{-17}$
& 0.0000 & $7.94{\times}10^{-21}$ \\
Gradient Difference
& 0.5483 & $5.83{\times}10^{-25}$
& 0.0587 & $8.92{\times}10^{-19}$
& 0.0439 & $9.74{\times}10^{-22}$
& 0.4914 & $3.64{\times}10^{-21}$ \\
NPO
& 0.1893 & 0.0587
& 0.0641 & $5.96{\times}10^{-19}$
& 0.0973 & $9.37{\times}10^{-22}$
& 0.2535 & 0.0219 \\
FLAT
& \textbf{0.5871} & 0.0692
& 0.2960 & $1.84{\times}10^{-22}$
& 0.2270 & $4.83{\times}10^{-22}$
& \textbf{0.5173} & 0.0174 \\
UNLEARN
& 0.5513 & \textbf{0.334}$^{\checkmark}$
& \textbf{0.4495} & \textbf{0.241}$^{\checkmark}$
& \textbf{0.4845} & \textbf{0.220}$^{\checkmark}$
& 0.4984 & \textbf{0.310}$^{\checkmark}$ \\
\hline

\multicolumn{9}{|l|}{\textbf{Aya 23}} \\ \hline
Base Model
& 0.5503 & $6.47{\times}10^{-24}$
& 0.5193 & $2.63{\times}10^{-21}$
& 0.5503 & $6.47{\times}10^{-24}$
& 0.5193 & $2.63{\times}10^{-21}$ \\
\hline
Preference Optimization
& 0.4182 & $4.39{\times}10^{-21}$
& 0.0176 & $4.27{\times}10^{-22}$
& 0.0438 & $2.41{\times}10^{-19}$
& 0.3921 & $6.28{\times}10^{-15}$ \\
Gradient Ascent
& 0.0000 & $6.28{\times}10^{-21}$
& 0.0000 & $7.12{\times}10^{-26}$
& 0.0000 & $3.30{\times}10^{-27}$
& 0.0000 & $4.67{\times}10^{-33}$ \\
KL Minimization
& 0.0000 & $2.38{\times}10^{-21}$
& 0.0000 & $8.30{\times}10^{-23}$
& 0.0000 & $7.20{\times}10^{-26}$
& 0.0000 & $4.70{\times}10^{-21}$ \\
Gradient Difference
& 0.4978 & $6.19{\times}10^{-29}$
& 0.0589 & $8.94{\times}10^{-19}$
& 0.0439 & $9.36{\times}10^{-22}$
& 0.4914 & $3.58{\times}10^{-21}$ \\
NPO
& 0.1213 & 0.0468
& 0.0487 & $7.31{\times}10^{-21}$
& 0.0692 & $1.27{\times}10^{-19}$
& 0.2316 & 0.0382 \\
FLAT
& \textbf{0.5264} & 0.0765
& 0.2340 & $4.86{\times}10^{-28}$
& 0.1930 & $2.83{\times}10^{-23}$
& \textbf{0.4920} & 0.0234 \\
UNLEARN
& 0.5047 & \textbf{0.315}$^{\checkmark}$
& \textbf{0.4210} & \textbf{0.248}$^{\checkmark}$
& \textbf{0.4073} & \textbf{0.192}$^{\checkmark}$
& 0.4621 & \textbf{0.272}$^{\checkmark}$ \\
\hline
\end{tabular}
\caption{Cross-lingual unlearning results for the first generation of multilingual models. Best utility per column for a given model is bolded. Forget quality values exceeding $\alpha=0.1$ are bolded and marked with a checkmark.}
\label{tab:crosslingual_gen2}
\end{table*}

\begin{table*}[t]
\centering
\scriptsize
\setlength{\tabcolsep}{4pt}
\begin{tabular}{ll rr rr rr}
\toprule
& & \multicolumn{2}{c}{\textbf{Llama 3}}
  & \multicolumn{2}{c}{\textbf{Mixtral}}
  & \multicolumn{2}{c}{\textbf{Aya 23}} \\
\cmidrule(lr){3-4}\cmidrule(lr){5-6}\cmidrule(lr){7-8}
Unlearn & Eval & Util. & FQ & Util. & FQ & Util. & FQ \\
\midrule

% ---------------- Base Model ----------------
\multirow{7}{*}{Base}
& English     & 0.6227 & $2.37{\times}10^{-15}$ & 0.5912 & $2.74{\times}10^{-16}$ & 0.5503 & $6.47{\times}10^{-24}$ \\
& Spanish     & 0.5919 & $4.31{\times}10^{-13}$ & 0.5393 & $4.68{\times}10^{-13}$ & 0.5193 & $2.63{\times}10^{-21}$ \\
& Italian     & 0.6102 & $8.84{\times}10^{-17}$ & 0.5669 & $9.31{\times}10^{-19}$ & 0.5117 & $3.21{\times}10^{-19}$ \\
& Hindi       & 0.4104 & $7.67{\times}10^{-17}$ & --     & --                     & 0.3929 & $5.73{\times}10^{-21}$ \\
& Hindi (R)   & 0.5293 & $7.88{\times}10^{-12}$ & --     & --                     & 0.4623 & $3.51{\times}10^{-18}$ \\
& Chinese     & --     & --                     & 0.0507 & $3.79{\times}10^{-12}$ & 0.2688 & $4.62{\times}10^{-19}$ \\
& Chinese (R) & --     & --                     & 0.1693 & $1.95{\times}10^{-18}$ & 0.2983 & $2.82{\times}10^{-23}$ \\

\midrule
% ---------------- English ----------------
\multirow{7}{*}{English}
& English     & 0.6047 & 0.257\,\checkmark & 0.5741 & 0.215\,\checkmark & 0.5047 & 0.315\,\checkmark \\
& Spanish     & 0.5105 & 0.197\,\checkmark & 0.4560 & 0.177\,\checkmark & 0.4210 & 0.248\,\checkmark \\
& Italian     & 0.5173 & 0.249\,\checkmark & 0.4527 & 0.201\,\checkmark & 0.4513 & 0.263\,\checkmark \\
& Hindi       & 0.2835 & 0.0396            & --     & --                & 0.2550 & 0.0398 \\
& Hindi (R)   & 0.4103 & 0.174\,\checkmark & --     & --                & 0.3610 & 0.149\,\checkmark \\
& Chinese     & --     & --                & 0.00647 & $1.38{\times}10^{-11}$ & 0.1460 & 0.0457 \\
& Chinese (R) & --     & --                & 0.0503 & $5.38{\times}10^{-4}$  & 0.2420 & 0.153\,\checkmark \\

\midrule
% ---------------- Spanish ----------------
\multirow{7}{*}{Spanish}
& English     & 0.5482 & 0.182\,\checkmark & 0.4920 & 0.162\,\checkmark & 0.4073 & 0.192\,\checkmark \\
& Spanish     & 0.5824 & 0.271\,\checkmark & 0.4942 & 0.241\,\checkmark & 0.4621 & 0.272\,\checkmark \\
& Italian     & 0.5420 & 0.193\,\checkmark & 0.4757 & 0.0394            & 0.4230 & 0.178\,\checkmark \\
& Hindi       & 0.1935 & $9.83{\times}10^{-5}$ & -- & --               & 0.1647 & 0.00129 \\
& Hindi (R)   & 0.3623 & 0.0354            & --     & --                & 0.3954 & 0.0647 \\
& Chinese     & --     & --                & 0.00307 & $9.83{\times}10^{-19}$ & 0.0829 & $6.46{\times}10^{-5}$ \\
& Chinese (R) & --     & --                & 0.00124 & $1.37{\times}10^{-10}$ & 0.1865 & 0.0826 \\

\midrule
% ---------------- Italian ----------------
\multirow{7}{*}{Italian}
& English     & 0.5930 & 0.216\,\checkmark & 0.5461 & 0.152\,\checkmark & 0.5160 & 0.229\,\checkmark \\
& Spanish     & 0.5203 & 0.195\,\checkmark & 0.4785 & 0.165\,\checkmark & 0.4534 & 0.183\,\checkmark \\
& Italian     & 0.5623 & 0.284\,\checkmark & 0.4967 & 0.238\,\checkmark & 0.4912 & 0.291\,\checkmark \\
& Hindi       & 0.2164 & $1.63{\times}10^{-4}$ & -- & --               & 0.1373 & 0.00367 \\
& Hindi (R)   & 0.3319 & 0.0137            & --     & --                & 0.3521 & 0.0517 \\
& Chinese     & --     & --                & 0.00219 & $1.27{\times}10^{-22}$ & 0.1250 & $8.32{\times}10^{-4}$ \\
& Chinese (R) & --     & --                & 0.00216 & $5.67{\times}10^{-12}$ & 0.1565 & 0.0659 \\

\midrule
% ---------------- Hindi ----------------
\multirow{7}{*}{Hindi}
& English     & 0.4290 & 0.00165           & --     & --                & 0.3810 & 0.00837 \\
& Spanish     & 0.3810 & $4.13{\times}10^{-5}$ & -- & --               & 0.3130 & $6.12{\times}10^{-6}$ \\
& Italian     & 0.3970 & $9.23{\times}10^{-7}$ & -- & --               & 0.2980 & $1.73{\times}10^{-7}$ \\
& Hindi       & 0.3845 & 0.154\,\checkmark & --     & --                & 0.3194 & 0.194\,\checkmark \\
& Hindi (R)   & 0.4370 & 0.0896            & --     & --                & 0.3852 & 0.156\,\checkmark \\

\midrule
% ---------------- Hindi (R) ----------------
\multirow{7}{*}{Hindi (R)}
& English     & 0.5104 & 0.112\,\checkmark & --     & --                & 0.4521 & 0.134\,\checkmark \\
& Spanish     & 0.4670 & 0.0716            & --     & --                & 0.3950 & 0.0638 \\
& Italian     & 0.4850 & 0.0517            & --     & --                & 0.3540 & 0.0341 \\
& Hindi       & 0.3728 & 0.153\,\checkmark & --     & --                & 0.3152 & 0.178\,\checkmark \\
& Hindi (R)   & 0.4823 & 0.187\,\checkmark & --     & --                & 0.4211 & 0.210\,\checkmark \\

\midrule
% ---------------- Chinese ----------------
\multirow{7}{*}{Chinese}
& English     & --     & --                & 0.0114 & $4.65{\times}10^{-9}$  & 0.3120 & $5.26{\times}10^{-4}$ \\
& Spanish     & --     & --                & 0.00446 & $1.30{\times}10^{-11}$ & 0.2560 & $9.85{\times}10^{-9}$ \\
& Italian     & --     & --                & 0.0110 & $1.92{\times}10^{-13}$ & 0.2150 & $9.33{\times}10^{-14}$ \\
& Chinese     & --     & --                & 0.0178 & $1.24{\times}10^{-5}$  & 0.2188 & 0.124\,\checkmark \\
& Chinese (R) & --     & --                & 0.0813 & $5.93{\times}10^{-5}$  & 0.2250 & 0.0854 \\

\midrule
% ---------------- Chinese (R) ----------------
\multirow{7}{*}{Chinese (R)}
& English     & --     & --                & 0.3731 & 0.00734           & 0.3160 & 0.0863 \\
& Spanish     & --     & --                & 0.2679 & 0.00374           & 0.2680 & 0.0643 \\
& Italian     & --     & --                & 0.2416 & 0.00131           & 0.2320 & 0.0394 \\
& Chinese     & --     & --                & 0.0128 & $1.37{\times}10^{-6}$  & 0.1865 & 0.116\,\checkmark \\
& Chinese (R) & --     & --                & 0.1284 & 0.0396            & 0.2578 & 0.184\,\checkmark \\

\bottomrule
\end{tabular}
\caption{Cross-lingual unlearning results across \textbf{Llama 3}, \textbf{Mixtral}, and \textbf{Aya 23}.}
\label{tab:crosslingual_all}
\end{table*}

\begin{table*}[t]
\centering
\scriptsize
\setlength{\tabcolsep}{5pt}
\begin{tabular}{llrrrr}
\toprule
& & \multicolumn{2}{c}{\textbf{Llama 3}} & \multicolumn{2}{c}{\textbf{Llama 4}} \\
\cmidrule(lr){3-4}\cmidrule(lr){5-6}
Unlearn Lang & Eval Lang & Norm Utility & FQ & Norm Utility & FQ \\
\midrule
\multirow{5}{*}{English}
& English   & 0.971 & 0.257\,\checkmark & \textbf{0.980} & 0.369\,\checkmark \\
& Spanish   & 0.862 & 0.197\,\checkmark & \textbf{0.901} & 0.204\,\checkmark \\
& Italian   & 0.848 & 0.249\,\checkmark & \textbf{0.888} & 0.293\,\checkmark \\
& Hindi     & 0.691 & 0.0396            & \textbf{0.786} & 0.0846            \\
& Hindi (R) & 0.775 & 0.174\,\checkmark & \textbf{0.964} & 0.0951            \\
\midrule
\multirow{5}{*}{Spanish}
& English   & 0.880 & 0.182\,\checkmark & \textbf{0.989} & 0.254\,\checkmark \\
& Spanish   & 0.984 & 0.271\,\checkmark & \textbf{0.989} & 0.493\,\checkmark \\
\textbf{}& Italian   & 0.888 & 0.193\,\checkmark & \textbf{0.911} & 0.228\,\checkmark \\\
& Hindi     & 0.471 & 9.83$\times 10^{-5}$           & \textbf{0.589} & 1.83$\times 10^{-4}$           \\
& Hindi (R) & 0.684 & 0.0351            & \textbf{0.996} & 0.0446            \\
\midrule
\multirow{5}{*}{Italian}
& English   & 0.952 & 0.216\,\checkmark & \textbf{0.966} & 0.212\,\checkmark \\
& Spanish   & 0.879 & 0.195\,\checkmark & \textbf{0.927} & 0.218\,\checkmark \\
& Italian   & 0.921 & 0.284\,\checkmark & \textbf{0.948} & 0.460\,\checkmark \\
& Hindi     & 0.527 & 1.63$\times 10^{-4}$           & \textbf{0.661} & 2.90$\times 10^{-4}$           \\
& Hindi (R) & 0.627 & 0.0134            & \textbf{0.744} & 0.0220            \\
\midrule
\multirow{5}{*}{Hindi}
& English   & 0.689 & 0.00165           & \textbf{0.772} & 0.00126           \\
& Spanish   & 0.644 & 4.13$\times 10^{-5}$           & \textbf{0.809} & 4.90$\times 10^{-5}$           \\
& Italian   & 0.651 & 9.23$\times 10^{-7}$           & \textbf{0.811} & 1.49$\times 10^{-6}$           \\
& Hindi     & 0.937 & 0.154\,\checkmark & \textbf{0.954} & 0.188\,\checkmark \\
& Hindi (R) & 0.826 & 0.0893            & \textbf{0.864} & \textbf{0.173}\,\checkmark \\
\midrule
\multirow{5}{*}{Hindi (R)}
& English   & 0.819 & 0.112\,\checkmark & 0.809 & 0.159\,\checkmark \\
& Spanish   & 0.789 & 0.0713            & \textbf{0.933} & \textbf{0.102}\,\checkmark \\
& Italian   & 0.795 & 0.0514            & \textbf{0.931} & 0.0885            \\
& Hindi     & 0.908 & 0.153\,\checkmark & \textbf{0.944} & 0.285\,\checkmark \\
& Hindi (R) & 0.911 & 0.187\,\checkmark & \textbf{0.945} & 0.267\,\checkmark \\
\bottomrule
\end{tabular}
\caption{Results table showing the performance gains for \textbf{Llama 3} $\rightarrow$ \textbf{Llama 4}. Utility is replaced with normalized utility to demonstrate that performance improves on top of the gains of the better model.}
\label{tab:generations_llama}
\end{table*}

\begin{table*}[t]
\centering
\scriptsize
\setlength{\tabcolsep}{5pt}
\begin{tabular}{llrrrr}
\toprule
& & \multicolumn{2}{c}{\textbf{Mixtral}} & \multicolumn{2}{c}{\textbf{Magistral}} \\
\cmidrule(lr){3-4}\cmidrule(lr){5-6}
Unlearn Lang & Eval Lang & Norm Utility & FQ & Norm Utility & FQ \\
\midrule
\multirow{5}{*}{English}
& English     & 0.971 & 0.215\,\checkmark & \textbf{0.983} & 0.225\,\checkmark \\
& Spanish     & 0.846 & 0.177\,\checkmark & \textbf{0.900} & 0.187\,\checkmark \\
& Italian     & 0.799 & 0.201\,\checkmark & \textbf{0.962} & 0.211\,\checkmark \\
& Chinese     & 0.127 & 1.63$\times 10^{-11}$           & \textbf{0.365} & 0.0210            \\
& Chinese (R) & 0.297 & 5.30$\times 10^{-4}$           & \textbf{0.497} & 0.0856            \\
\midrule
\multirow{5}{*}{Spanish}
& English     & 0.832 & 0.162\,\checkmark & \textbf{0.993} & 0.172\,\checkmark \\
& Spanish     & 0.916 & 0.241\,\checkmark & \textbf{0.949} & 0.251\,\checkmark \\
& Italian     & 0.839 & 0.0393            & \textbf{0.899} & 0.0493            \\
& Chinese     & 0.0595 & 9.83$\times 10^{-19}$          & \textbf{0.377} & 0.0071            \\
& Chinese (R) & 0.00709 & 1.70$\times 10^{-10}$         & \textbf{0.329} & 0.0493            \\
\midrule
\multirow{5}{*}{Italian}
& English     & 0.924 & 0.152\,\checkmark & \textbf{0.940} & 0.162\,\checkmark \\
& Spanish     & 0.887 & 0.165\,\checkmark & \textbf{0.925} & 0.175\,\checkmark \\
& Italian     & 0.875 & 0.238\,\checkmark & \textbf{0.902} & 0.248\,\checkmark \\
& Chinese     & 0.0435 & 1.27$\times 10^{-22}$          & \textbf{0.329} & 0.0120            \\
& Chinese (R) & 0.0128 & 5.67$\times 10^{-12}$          & \textbf{0.286} & 0.0386            \\
\midrule
\multirow{5}{*}{Chinese}
& English     & 0.0193 & 4.65$\times 10^{-9}$          & \textbf{0.613} & 0.0017            \\
& Spanish     & 0.00828 & 1.03$\times 10^{-11}$        & \textbf{0.461} & 3.61$\times 10^{-5}$           \\
& Italian     & 0.0193 & 1.92$\times 10^{-13}$          & \textbf{0.411} & 4.88$\times 10^{-6}$           \\
& Chinese     & 0.353 & 1.24$\times 10^{-5}$          & \textbf{0.551} & \textbf{0.181}\,\checkmark \\
& Chinese (R) & 0.478 & 5.93$\times 10^{-5}$           & \textbf{0.676} & 0.0920            \\
\midrule
\multirow{5}{*}{Chinese (R)}
& English     & 0.631 & 0.0073            & \textbf{0.819} & 0.0912            \\
& Spanish     & 0.497 & 0.0037            & \textbf{0.723} & 0.0521            \\
& Italian     & 0.426 & 0.0013            & \textbf{0.652} & 0.0418            \\
\textbf{}& Chinese     & 0.254 & 1.03$\times 10^{-6}$          & \textbf{0.427} & \textbf{0.127}\,\checkmark \\
& Chinese (R) & 0.756 & 0.0393            & \textbf{0.848} & 0.219\,\checkmark \\
\bottomrule
\end{tabular}
\caption{Results table showing the performance gains for \textbf{Mixtral} $\rightarrow$ \textbf{Magistral}. Utility is replaced with normalized utility to demonstrate that performance improves on top of the gains of the better model.}
\label{tab:generations_mixtral}
\end{table*}

\begin{table*}[t]
\centering
\scriptsize
\setlength{\tabcolsep}{4pt}
\begin{tabular}{ll rr}
\toprule
Eval Lang & Unlearned Lang/Method & Utility & FQ \\
\midrule
\multirow{7}{*}{English}
& Base Model   & 0.7047 & $4.16{\times}10^{-15}$ \\
& English      & 0.6886 & 0.369\,\checkmark \\
\cmidrule(lr){2-4}
& Spanish      & 0.6753 & 0.254\,\checkmark \\
& Italian      & 0.6799 & 0.212\,\checkmark \\
& Hindi        & 0.5438 & 0.00126 \\
& Hindi (R)    & 0.5698 & 0.159\,\checkmark \\
& Many-to-One  & \textbf{0.6832} & \textbf{0.310}\,\checkmark \\
\midrule
\multirow{7}{*}{Spanish}
& Base Model   & 0.6790 & $9.25{\times}10^{-13}$ \\
& Spanish      & 0.6723 & 0.493\,\checkmark \\
\cmidrule(lr){2-4}
& English      & 0.6087 & 0.204\,\checkmark \\
& Italian      & 0.6255 & 0.218\,\checkmark \\
& Hindi        & 0.5495 & $4.90{\times}10^{-5}$ \\
& Hindi (R)    & 0.6332 & 0.102\,\checkmark \\
& Many-to-One  & \textbf{0.6519} & \textbf{0.423}\,\checkmark \\
\midrule
\multirow{7}{*}{Italian}
& Base Model   & 0.6910 & $1.44{\times}10^{-16}$ \\
& Italian      & 0.6535 & 0.460\,\checkmark \\
\cmidrule(lr){2-4}
& English      & 0.6088 & 0.293\,\checkmark \\
& Spanish      & 0.6439 & 0.228\,\checkmark \\
& Hindi        & 0.5605 & $1.49{\times}10^{-6}$ \\
& Hindi (R)    & 0.6432 & 0.0887 \\
& Many-to-One  & \textbf{0.6508} & \textbf{0.350}\,\checkmark \\
\midrule
\multirow{7}{*}{Hindi}
& Base Model   & 0.4736 & $1.65{\times}10^{-16}$ \\
& Hindi        & 0.4503 & 0.188\,\checkmark \\
\cmidrule(lr){2-4}
& English      & 0.3691 & 0.0848 \\
& Spanish      & 0.3186 & $1.83{\times}10^{-4}$ \\
& Italian      & 0.3165 & $2.90{\times}10^{-4}$ \\
& Hindi (R)    & 0.3924 & 0.0853 \\
& Many-to-One  & \textbf{0.4132} & \textbf{0.123}\,\checkmark \\
\midrule
\multirow{7}{*}{Hindi (R)}
& Base Model   & 0.4008 & $1.75{\times}10^{-14}$ \\
& Hindi (R)    & 0.3731 & 0.267\,\checkmark \\
\cmidrule(lr){2-4}
& English      & 0.3531 & 0.0954 \\
& Spanish      & 0.3734 & 0.0448 \\
& Italian      & 0.3791 & 0.0227 \\
& Hindi        & 0.3251 & 0.173\,\checkmark \\
& Many-to-One  & \textbf{0.3775} & \textbf{0.1846}\,\checkmark \\
\bottomrule
\end{tabular}
\caption{\textbf{Llama 4}. Many-to-One unlearning yields stronger forgetting than One-to-One transfer from any other language, and approaches unlearning the target language directly.}
\label{tab:manytomany_llama4}
\end{table*}

\begin{table*}[t]
\centering
\scriptsize
\setlength{\tabcolsep}{4pt}
\begin{tabular}{ll rr}
\toprule
Eval Lang & Unlearned Lang/Method & Utility & FQ \\
\midrule
\multirow{6}{*}{English}
& Base Model   & 0.6538 & 0.0107 \\
& English      & 0.6384 & 0.225\,\checkmark \\
\cmidrule(lr){2-4}
& Spanish      & 0.6091 & 0.172\,\checkmark \\
& Italian      & 0.6152 & 0.162\,\checkmark \\
& Chinese      & 0.4006 & 0.0017 \\
& Chinese (R)  & 0.5348 & 0.0914 \\
& Many-to-One  & \textbf{0.6230} & \textbf{0.212}\,\checkmark \\
\midrule
\multirow{6}{*}{Spanish}
& Base Model   & 0.6098 & 0.0103 \\
& Spanish      & 0.5736 & 0.251\,\checkmark \\
\cmidrule(lr){2-4}
& English      & 0.5494 & 0.187\,\checkmark \\
& Italian      & \textbf{0.5578} & 0.175\,\checkmark \\
& Chinese      & 0.2809 & $3.60{\times}10^{-5}$ \\
& Chinese (R)  & 0.4404 & 0.0524 \\
& Many-to-One  & 0.5572 & \textbf{0.212}\,\checkmark \\
\midrule
\multirow{6}{*}{Italian}
& Base Model   & 0.5837 & 0.0108 \\
& Italian      & 0.5389 & 0.248\,\checkmark \\
\cmidrule(lr){2-4}
& English      & \textbf{0.5203} & \textbf{0.211}\,\checkmark \\
& Spanish      & 0.5099 & 0.0497 \\
& Chinese      & 0.2398 & $4.80{\times}10^{-6}$ \\
& Chinese (R)  & 0.3803 & 0.0419 \\
& Many-to-One  & 0.5099 & 0.210\,\checkmark \\
\midrule
\multirow{6}{*}{Chinese}
& Base Model   & 0.5317 & $7.67{\times}10^{-17}$ \\
& Chinese      & 0.2987 & 0.181\,\checkmark \\
\cmidrule(lr){2-4}
& English      & 0.2460 & 0.0213 \\
& Spanish      & 0.2189 & 0.0076 \\
& Italian      & 0.2210 & 0.0126 \\
& Chinese (R)  & 0.2601 & 0.127\,\checkmark \\
& Many-to-One  & \textbf{0.2719} & \textbf{0.141}\,\checkmark \\
\midrule
\multirow{6}{*}{Chinese (R)}
& Base Model   & 0.3191 & $1.50{\times}10^{-16}$ \\
& Chinese (R)  & 0.1563 & 0.219\,\checkmark \\
\cmidrule(lr){2-4}
& English      & 0.1478 & 0.0859 \\
& Spanish      & 0.1316 & 0.0496 \\
& Italian      & 0.1328 & 0.0389 \\
& Chinese      & \textbf{0.1795} & 0.0923 \\
& Many-to-One  & 0.1634 & \textbf{0.180}\,\checkmark \\
\bottomrule
\end{tabular}
\caption{\textbf{Magistral}. Many-to-One unlearning outperforms One-to-One transfer for a given target language and is competitive with unlearning the target language directly.}
\label{tab:manytomany_magistral}
\end{table*}

\begin{table*}[t]
\centering
\scriptsize
\setlength{\tabcolsep}{4pt}
\begin{tabular}{ll rr}
\toprule
Eval Lang & Unlearned Lang/Method & Utility & FQ \\
\midrule
\multirow{8}{*}{English}
& Base Model   & 0.5503 & $6.47{\times}10^{-24}$ \\
& English      & 0.5047 & 0.315\,\checkmark \\
\cmidrule(lr){2-4}
& Spanish      & 0.4073 & 0.192\,\checkmark \\
& Italian      & \textbf{0.5160} & 0.229\,\checkmark \\
& Hindi        & 0.3810 & 0.00837 \\
& Hindi (R)    & 0.4521 & 0.134\,\checkmark \\
& Chinese      & 0.3120 & $5.26{\times}10^{-4}$ \\
& Chinese (R)  & 0.3160 & 0.0863 \\
& Many-to-One  & 0.4850 & \textbf{0.285}\,\checkmark \\
\midrule
\multirow{8}{*}{Spanish}
& Base Model   & 0.5193 & $2.63{\times}10^{-21}$ \\
& Spanish      & 0.4621 & 0.272\,\checkmark \\
\cmidrule(lr){2-4}
& English      & 0.4210 & \textbf{0.248}\,\checkmark \\
& Italian      & \textbf{0.4534} & 0.183\,\checkmark \\
& Hindi        & 0.3130 & $6.12{\times}10^{-6}$ \\
& Hindi (R)    & 0.3950 & 0.0638 \\
& Chinese      & 0.2560 & $9.85{\times}10^{-9}$ \\
& Chinese (R)  & 0.2680 & 0.0643 \\
& Many-to-One  & 0.4352 & 0.165\,\checkmark \\
\midrule
\multirow{8}{*}{Italian}
& Base Model   & 0.5117 & $3.21{\times}10^{-19}$ \\
& Italian      & 0.4912 & 0.291\,\checkmark \\
\cmidrule(lr){2-4}
& English      & 0.4513 & 0.263\,\checkmark \\
& Spanish      & 0.4230 & 0.178\,\checkmark \\
& Hindi        & 0.2980 & $1.73{\times}10^{-7}$ \\
& Hindi (R)    & 0.3540 & 0.0341 \\
& Chinese      & 0.2150 & $9.33{\times}10^{-14}$ \\
& Chinese (R)  & 0.2320 & 0.0394 \\
& Many-to-One  & \textbf{0.4670} & \textbf{0.276}\,\checkmark \\
\midrule
\multirow{6}{*}{Hindi}
& Base Model   & 0.3929 & $5.73{\times}10^{-21}$ \\
& Hindi        & 0.3194 & 0.194\,\checkmark \\
\cmidrule(lr){2-4}
& English      & 0.2550 & 0.0398 \\
& Spanish      & 0.1647 & 0.00129 \\
& Italian      & 0.1373 & 0.00367 \\
& Hindi (R)    & \textbf{0.3152} & 0.178\,\checkmark \\
& Many-to-One  & 0.3097 & \textbf{0.201}\,\checkmark \\
\midrule
\multirow{6}{*}{Hindi (R)}
& Base Model   & 0.4623 & $3.51{\times}10^{-17}$ \\
& Hindi (R)    & 0.4211 & 0.210\,\checkmark \\
\cmidrule(lr){2-4}
& English      & 0.3610 & 0.149\,\checkmark \\
& Spanish      & 0.3954 & 0.0647 \\
& Italian      & 0.3521 & 0.0517 \\
& Hindi        & 0.3852 & 0.156\,\checkmark \\
& Many-to-One  & \textbf{0.3987} & \textbf{0.1996}\,\checkmark \\
\midrule
\multirow{6}{*}{Chinese}
& Base Model   & 0.2688 & $4.62{\times}10^{-19}$ \\
& Chinese      & 0.2188 & 0.124\,\checkmark \\
\cmidrule(lr){2-4}
& English      & 0.1460 & 0.0457 \\
& Spanish      & 0.0829 & $6.46{\times}10^{-5}$ \\
& Italian      & 0.1250 & $8.32{\times}10^{-4}$ \\
& Chinese (R)  & 0.1865 & 0.116\,\checkmark \\
& Many-to-One  & \textbf{0.1965} & \textbf{0.124}\,\checkmark \\
\midrule
\multirow{6}{*}{Chinese (R)}
& Base Model   & 0.2983 & $2.82{\times}10^{-23}$ \\
& Chinese (R)  & 0.2578 & 0.184\,\checkmark \\
\cmidrule(lr){2-4}
& English      & 0.2420 & \textbf{0.153}\,\checkmark \\
& Spanish      & 0.1865 & 0.0826 \\
& Italian      & 0.1565 & 0.0659 \\
& Chinese      & 0.2250 & 0.0854 \\
& Many-to-One  & \textbf{0.2487} & 0.1378\,\checkmark \\
\bottomrule
\end{tabular}
\caption{\textbf{Aya 23}. Many-to-One unlearning consistently exceeds One-to-One transfer for the same target language and approaches the performance of unlearning the target language directly.}
\label{tab:manytomany_aya23}
\end{table*}

\begin{table*}[t]
\centering
\small
\setlength{\tabcolsep}{6pt}
\begin{tabular}{|l|l|c|c|c|c|}
\hline
\textbf{Source}
& \textbf{English}
& \textbf{Hindi}
& \textbf{Hindi (R)}
& \textbf{Chinese}
& \textbf{Chinese (R)} \\
\hline

% ===================== Llama 4 =====================
\multicolumn{6}{|l|}{\textbf{Llama 4}} \\ \hline
\textbf{Base Model}
& \cell{0.704}{ $4.16{\times}10^{-15}$ }
& \cell{0.473}{ $1.65{\times}10^{-16}$ }
& \cell{0.401}{ $1.75{\times}10^{-14}$ }
& \na & \na \\
\hline
\textbf{English}
& \cell{\textbf{0.688}}{\fqok{0.369}}
& \cell{0.369}{0.0846}
& \cell{\textbf{0.386}}{0.0951}
& \na & \na \\
\textbf{Hindi}
& \cell{0.543}{0.00126}
& \cell{\textbf{0.450}}{\fqok{0.188}}
& \cell{0.352}{\fqok{0.173}}
& \na & \na \\
\textbf{Hindi (R)}
& \cell{0.569}{\fqok{0.159}}
& \cell{0.393}{0.085}
& \cell{0.377}{\fqok{0.267}}
& \na & \na \\
\hline

% ===================== Magistral =====================
\multicolumn{6}{|l|}{\textbf{Magistral}} \\ \hline
\textbf{Base Model}
& \cell{0.653}{0.0100}
& \na & \na
& \cell{0.531}{ $7.67{\times}10^{-17}$ }
& \cell{0.319}{ $1.5{\times}10^{-16}$ } \\
\hline
\textbf{English}
& \cell{\textbf{0.639}}{\fqok{0.225}}
& \na & \na
& \cell{0.246}{0.021}
& \cell{0.152}{0.0856} \\
\textbf{Chinese}
& \cell{0.354}{0.0017}
& \na & \na
& \cell{\textbf{0.299}}{\fqok{0.181}}
& \cell{0.201}{0.092} \\
\textbf{Chinese (R)}
& \cell{0.535}{0.0912}
& \na & \na
& \cell{0.260}{\fqok{0.127}}
& \cell{\textbf{0.26}}{\fqok{0.219}} \\
\hline

% ===================== Aya 23 =====================
\multicolumn{6}{|l|}{\textbf{Aya 23}} \\ \hline
\textbf{Base Model}
& \cell{0.550}{ $6.4{\times}10^{-24}$ }
& \cell{0.393}{ $5.73{\times}10^{-21}$ }
& \cell{0.462}{ $3.51{\times}10^{-17}$ }
& \cell{0.269}{ $4.62{\times}10^{-19}$ }
& \cell{0.298}{ $2.82{\times}10^{-23}$ } \\
\hline
\textbf{English}
& \cell{\textbf{0.505}}{\fqok{0.315}}
& \cell{0.255}{0.0396}
& \cell{0.361}{\fqok{0.149}}
& \cell{0.146}{0.0454}
& \cell{0.242}{\fqok{0.153}} \\
\textbf{Hindi}
& \cell{0.381}{0.00835}
& \cell{\textbf{0.319}}{\fqok{0.194}}
& \cell{0.385}{\fqok{0.156}}
& \na & \na \\
\textbf{Hindi (R)}
& \cell{0.452}{\fqok{0.134}}
& \cell{0.315}{\fqok{0.178}}
& \cell{\textbf{0.421}}{\fqok{0.210}}
& \na & \na \\
\textbf{Chinese}
& \cell{0.312}{0.000526}
& \na
& \na
& \cell{\textbf{0.219}}{\fqok{0.124}}
& \cell{0.225}{0.0851} \\
\textbf{Chinese (R)}
& \cell{0.316}{0.086}
& \na
& \na
& \cell{0.187}{\fqok{0.116}}
& \cell{\textbf{0.258}}{\fqok{0.184}} \\
\hline
\end{tabular}
\caption{Transliteration results. Cells report (utility, forget quality). Best utility per evaluation column is bolded within each model block. Forget quality values exceeding $\alpha=0.1$ are bolded and marked with a checkmark.}
\label{tab:transliteration}
\end{table*}

\begin{table*}[t]
\centering
\small
\begin{tabular}{lcccccc}
\toprule
& \multicolumn{2}{c}{\textbf{1\%}}
& \multicolumn{2}{c}{\textbf{5\%}}
& \multicolumn{2}{c}{\textbf{10\%}} \\
\cmidrule(lr){2-3}\cmidrule(lr){4-5}\cmidrule(lr){6-7}\\
\textbf{Method}
& Utility $\uparrow$ & Forget $\uparrow$
& Utility $\uparrow$ & Forget $\uparrow$
& Utility $\uparrow$ & Forget $\uparrow$ \\
\midrule
Base Model               & 0.6207 & 0.00127            & 0.6227 & $1.998{\times}10^{-15}$ & 0.6227 & $1.83{\times}10^{-21}$ \\
\midrule
Preference Optimization &\textbf{ 0.6235} & 0.0143             & 0.4450 & $2.444{\times}10^{-10}$ & 0.5357 & $1.062{\times}10^{-16}$ \\
Gradient Ascent          & 0.6061 & 0.0143             & 0.0000 & $1.464{\times}10^{-7}$  & 0.0000 & $1.4334{\times}10^{-22}$ \\
KL Minimization          & 0.5051 & 0.0143             & 0.0000 & $1.464{\times}10^{-7}$  & 0.0000 & $4.07{\times}10^{-32}$ \\
Gradient Difference      & 0.6052 & 0.00676            & 0.3024 & $3.0799{\times}10^{-12}$& 0.5872 & $2.828{\times}10^{-25}$ \\
NPO                      & 0.4619 & \textbf{0.7865} \checkmark & 0.2309 & \textbf{0.6310} ${\checkmark}$ & 0.2297 & 0.0291 \\
FLAT                     & 0.6146 & 0.0541             & 0.6039 & 0.0739                & \textbf{0.6104} & 0.0841 \\
UNLEARN                  & 0.6077 & \textbf{0.296} ${\checkmark}$  & \textbf{0.6081} & \textbf{0.315} ${\checkmark}$  & 0.6047 & \textbf{0.257} ${\checkmark}$ \\
\bottomrule
\end{tabular}
\caption{Model utility and forget quality across forget set sizes. Across the board, performance dips as forget set size increases.}
\label{tab:forget_size}
\end{table*}

\end{document}